\documentclass{article}
\usepackage{authblk}
\usepackage{xurl}
\usepackage{hyperref}

\usepackage{abstract}
\usepackage{graphicx}
\usepackage{rotating}
\usepackage{tcolorbox}

\usepackage{mathptmx}  
\usepackage{sectsty}   
\usepackage{fancyhdr}  

\usepackage{titlesec}
\def\shorttitle{Unlocking the Archives}

\titleformat{\section}{\fontsize{12}{14}\selectfont\bfseries}{}{0pt}{}
\titleformat{\subsection}{\fontsize{12}{14}\selectfont\bfseries}{}{0pt}{}

\makeatletter
\def\@maketitle{%
  \newpage
  \null
  \vskip 1em%
  \begin{center}%
    \hrule height 0.5pt \vspace{2em}%
    \let \footnote \thanks
    {\LARGE \@title \par}%
    \vspace{2em}\hrule height 0.5pt%
    \vskip 1.5em%
    {\small \@author \par}%
  \end{center}%
  \par
  \vskip 1.8em}
\makeatother

\vspace{2em}  

\fancypagestyle{plain}{%
  \fancyhf{}  
}

\AtBeginDocument{
  \thispagestyle{plain}
  \global\pagenumbering{gobble}  
  \clearpage
  \global\pagenumbering{arabic}  
}  

\title{Unlocking the Archives: Large Language Models Achieve State-of-the-Art Performance on the Transcription of Handwritten Historical Documents}
\date{}

\author[1]{Mark Humphries\thanks{Corresponding author: mhumphries@wlu.ca}}
\author[1]{Lianne C. Leddy}
\author[1]{Quinn Downton}
\author[1]{Meredith Legace}
\author[1]{John McConnell}
\author[1]{Isabella Murray}
\author[1]{Elizabeth Spence}

\affil[1]{\textit{Wilfrid Laurier University}}

\begin{document}

\maketitle

\begin{abstract}
This study demonstrates that Large Language Models (LLMs) can transcribe historical handwritten documents with significantly higher accuracy than specialized Handwritten Text Recognition (HTR) software, while being faster and more cost-effective. We introduce an open-source software tool called \textit{Transcription Pearl} that leverages these capabilities to automatically transcribe and correct batches of handwritten documents using commercially available multimodal LLMs from OpenAI, Anthropic, and Google. In tests on a diverse corpus of 18th/19th century English language handwritten documents, LLMs achieved Character Error Rates (CER) of 5.7 to 7\% and Word Error Rates (WER) of 8.9 to 15.9\%, improvements of 14\% and 32\% respectively over specialized state-of-the-art HTR software like \textit{Transkribus}. Most significantly, when LLMs were then used to correct those transcriptions as well as texts generated by conventional HTR software, they achieved near-human levels of accuracy, that is CERs as low as 1.8\% and WERs of 3.5\%. The LLMs also completed these tasks 50 times faster and at approximately 1/50th the cost of proprietary HTR programs. These results demonstrate that when LLMs are incorporated into software tools like \textit{Transcription Pearl}, they provide an accessible, fast, and highly accurate method for mass transcription of historical handwritten documents, significantly streamlining the digitization process.
\end{abstract}

 \section{Introduction}
 
 As historians and archivists digitize ever larger collections of handwritten records, accurate transcription remains a barrier to their systematic analysis, publication, and accessibility. It is also the most time-consuming and costly aspect of the digitization process. The field of Handwritten Text Recognition (HTR) attempts to automate the task through machine learning but it is a complex and technologically advanced subfield of computer vision and artificial intelligence research, an area in which few historians are comfortable operating. 

In recent years, programs like as \textit{Transkribus}, run by the READ-COOP cooperative which includes more than 150 major universities and archives, has made HTR far more accessible through graphical, “drag and drop” interfaces (Kahle et al 2017; Nockels et al 2022). \textit{Transkribus} largely automates the most complex image pre-processing and segmentation elements of the workflow and allows users to use pretrained neural network models to generate a rough transcription of handwritten documents. These models are highly accurate when they have been trained to recognize a specific individual’s handwriting, achieving word-level accuracy between 85 and 95\% (Al Kendi, 2024). That said, they are also moderately expensive and time-consuming to use at roughly \$0.27 USD per page with variable turnaround times. Although actual processing normally takes 15-20 minutes, depending on customer volumes transcription requests can be placed in a queue for a day or more.

A more important issue is that because handwriting is unique, the most impressive results do not normally reflect most users’ experiences with HTR. Historians are often discouraged when they find that most HTR models have significant difficulty generalizing “out of the box” to new hands, document formats, or styles that they did not encounter in training. By “out-of-the-box”, we mean a model that can be employed “as is” without the user having to train the model on a specific handwriting style. On such tasks, they typically achieve word level accuracy of only 50-75\% which make the results largely unusable without major editing. Teaching or fine-tuning models to recognize a specific, individual handwriting style requires users to generate perfect transcriptions of \textit{at least} 75 handwritten pages (about 15,000 words)—referred to as “ground-truth” documents—which can then be used as training data. As a result, unless historians are working with hundreds or thousands of pages of documents written in a single hand, this fine-tuning process is rarely worth the effort. This means that for most historians, HTR remains a tantalizing but impractical solution to the transcription problem.

In this paper, we introduce \textit{Transcription Pearl}, an easy-to-use software tool that automatically transcribes batches of handwritten documents “out-of-the-box” using widely available multi-modal Large Language Models (LLMs) like GPT-4o, Claude Sonnet-3.5, and Gemini 1.5-Pro, quickly, cost effectively, and to a high degree of accuracy. Depending on how capitalization, punctuation, and historical spelling errors are evaluated, it achieves accuracy levels of between 84 and 93\% on transcription tasks without the need for image pre-processing or fine-tuning. We also demonstrate that unlike conventional HTR software, LLMs can be used to correct transcriptions generated by other LLMs as well as \textit{Transkribus} to levels of accuracy that would normally require extensive fine-tuning, that is above 96\%. At this level of accuracy these transcriptions not only approach human levels of accuracy but are “good enough” for most day-to-day use cases such as full text keyword and semantic search as well as improving accessibility and readability. We thus show that the application of LLMs to HTR potentially provides an accessible, affordable, fast, and accurate pathway for both mass transcription projects and individual historians working on their own corpus of records.

\section{Human to Automated Transcription}

Over the last two decades, most historians have begun to incorporate digital photography into their archival research, amassing large personal collections of documents. In the “shooting the archives” approach, historians capture all or most of the documents in a given file or volume in the archives (Keeling and Sandlos, 2011). This approach offers several clear advantages over older analog approaches. It not only saves time and expense in the field, but also allows for a more comprehensive examination of the material. In the past, when most historians relied on notes and carefully selected photocopies, their ability to re-examine evidence based on new information was relatively limited, often necessitating multiple trips to the same archive. Now it is essentially unlimited: digital documents can always be re-examined or reassessed based on new information or context. Yet this creates its own issues as the volume of archival materials that historians amass are immense. In an ideal world, these records would be made full text searchable by keyword or semantic meaning, but that requires transcription. 

Transcription is a necessary precursor to any digital manipulation of handwritten sources. Yet it is a time-consuming and expensive process. Professional transcription services now rarely offer text-to-text services, but those that do can charge between \$6 and \$12 USD per page, offering turn-around times varying from next day service to more than two weeks (GMR Transcription, 2024). Historians more commonly employ student research assistants on these tasks and while we could not locate any standard “turn-around-times” in the literature, in our experience student RAs usually transcribe around 5-7 pages per day. For this reason, advancements in the automated transcription of handwritten text are of significant interest to practicing historians as well as archives looking to make their holding more accessible (Romein et al 2020). 

The automatic recognition of printed and handwritten text has been an important area of research in computer vision and pattern recognition since the early 1960s (Wang 2023). The first Optical Character Recognition (OCR) systems focused on printed text, utilizing template matching and pattern recognition to identify characters in scanned images. These systems required high-quality inputs and struggled with variations in fonts and layouts, limiting their effectiveness (Memon et al., 2020). By the late 1990s, OCR systems that relied on machine learning approaches and visual pattern matching met with greater success, resulting in the availability of reliable commercial OCR software like ABBYY FineReader and the open-source Tesseract by the mid 2000s (Tafti 2016).

 Handwritten Text Recognition (HTR) presents a more complex problem compared with printed text due to the vast diversity in individual handwriting styles, cursive scripts, and inconsistent character shapes and spacing (Wigington 2018). Traditional HTR methods employed statistical modelling combined with extensive preprocessing steps—including image binarization, noise reduction, and segmentation into lines or words—to standardize inputs for recognition algorithms (Plamondon \& Srihari 2000). Later Convolutional Neural Networks (CNNs) were trained to capture features like strokes and edges (LeCun, et al., 1998), while Recurrent Neural Networks (RNNs), particularly Long Short-Term Memory (LSTM) networks, modeled the sequential nature of handwriting (Graves et al 2008). Hybrid models combining CNNs and LSTMs enabled end-to-end learning from raw images to transcribed text, simplifying the process by reducing the need for manual feature extraction (Bluche 2016). The adoption of the transformer architecture and attention mechanisms further enhanced performance by allowing models to focus on the most relevant parts of the input when generating transcriptions (Cheng et al 2017; d’Arce 2022).

In a typical HTR workflow, documents are digitized then the resulting images are preprocessed to better allow HTR software to “read” the text. This can be a technically complicated process that involves noise removal, de-skewing (straightening), resizing, and binarization (making the images black and white) to standardize and improve images for processing (Sanchez-DelaCruz 2024; Al Kendi 2024). Next the text is segmented into lines, words, and characters, either manually or by a programs that try to identify those components on the page (Wigington 2018; Al Kendi 2024). Finally, features are extracted from the image using neural networks trained to identify visual patterns in the structures of the handwritten text and its textures as well as statistical information about the distribution of pixels, gradients, and zoning. Classification systems of various designs match the features extracted from the image with predefined patterns learned in training to generate outputs (the transcription). These pre-processed images are then sent to the neural network model which generates a transcription (d’Arce 2022; Al Kendi 2024). The goal is to produce a transcript with as few errors as possible.

\section{Measuring Errors}

In the field of HTR, accuracy is measured using Character Error Rates (CER) and Word Error Rates (WER), the latter normally being roughly 3 to 4 times more than the former (Sánchez et al 2019). Error rates are established by summing the substitutions, additions, and deletions in a given document and dividing them by the total characters or words in the document. Here the ideal is a perfect, proof-read, error-free text, yet in most real-world scenarios, the transcriptions generated by historians and their research assistants would rarely achieve to this standard. The true alternative to an HTR generated document is usually a human non-expert generated transcription that would rarely (if ever) be “error free”. Even professional human transcription services only guarantee 99\% accuracy, meaning 1 word in 100 is expected to be incorrectly transcribed—a rate that comes with the important caveat that the image, style, and handwriting must be clear, which is not always the case with historical documents. 

There is comparatively little information available on non-expert human error rates. For the transcription of speech, a typically cited WER is between 4 and 10\% (Stolcke 2017). Less work has been done on textual transcription, but studies in the transcription of patient medical records in clinical settings show similar error rates of around 10\% (Feng et al, 2020; Nordo et al 2017). Only one study could be located focusing specifically on human error rates in the transcription of historical documents—albeit on early modern Italian manuscripts written in an uncommon script—and it revealed WERs as high as 35-43\% depending on how errors were evaluated (Oliveira, 2018).

One of the primary issues in evaluating the usefulness of historical transcription is a lack of commonly agreed upon standards. While oral historians have written extensively and thoughtfully about best practices for transcribing interviews, few have given the same level of consideration to textual transcription (Bergen 2019; Leddy 2010; Strong 2018). In one of the few scholarly studies in the area, Stevens and Burg argue that transcription is always a “form of translation and requires editors to make innumerable decisions about how to present documents” (1997, 71). On one extreme, these include visual choices related to standardizing layouts, things like reordering addresses, salutations, datelines, marginalia, and paragraph indentations. But they also include emendations, that is changes made to transcriptions to improve readability and interpretability such as standardizing or modernizing capitalization, punctuation, and spelling; emendations can also include more intrusive acts like writing out abbreviations and supplying missing letters and words (Stevens and Burg 1997, 72). In place of shared standards, historians “use different types of editorial apparatus and different degrees of emendation depending on the nature of the documents and the intended audience” (Stevens and Burg 1997, 72). 

Measuring WER and CER in automatically transcribed documents is especially difficult with historical documents (Vidal et al 2023). For example, a handwritten text might be accurately rendered by HTR, but reassembled out of order due to the mis-sequencing of text lines. At the same time, ambiguous punctuation, capital letters, and non alphabetical character formation can also artificially inflate error rates (Perdiki 2023). While strict attention to capitalization and punctuation can often be important for publication (and even then, can be highly subjective), it is less important if the goal is simply to allow someone to accurately navigate a large corpus of documents via full text search.

Beyond questions about emendation and transcription standards, working with digitized historical documents poses a number of other unique challenges for HTR, specifically issues like paper degradation, ink fading, and archaic scripts as well as the prevalence of poorly scanned microfilm and microfiche records (Sánchez et al 2019). Many also include marginalia, insertions, strikeouts, and other artifacts that convey important semantic content but are often difficult to decipher and represent, even for expert humans. At the same time, many (if not most) of the document images used by researchers in real-world scenarios were taken with handheld cameras (often phone cameras), in poor lighting, and with a wide range of clarity, focal lengths, resolutions, and colour profiles (Keeling et al 2011). In combination, these factors make it exceedingly difficult to develop generalized models capable of handling the range of diverse inputs typical of real-world use-cases (Christlein et al 2018).

\section{Minimizing HTR Errors with Fine Tuning}

\textit{Transkribus} is the most popular HTR tool with historical researchers (Kahle et al 2017). Developed by the READ-COOP cooperative in the mid 2010s, the program uses neural network models with various architectures all trained and fine-tuned via conventional supervised machine learning techniques. In this approach, training data consists of an image of a handwritten document and a corresponding ground-truth transcription. While \textit{Transkribus} reports CERs of around 4\% on its validation sets (that is, subsets of the training data held back from the model for testing), the models have difficulty generalizing to hands that were not represented in their training data. On these “out of distribution” tasks, the performance of \textit{Transkribus} base models is highly variable, ranging from CERs of 8-25\% and WERs of 15-50\%, often limiting the usefulness of the transcribed text for most applications (Nihart 2022). The results vary so widely because handwriting styles are unique to each individual. Where a given style is similar to one that is well-represented in the training data, the model will typically perform better; when a hand is dissimilar from those in the training data, it will not perform as well. 

The most common solution to problems of accuracy is to recommend fine tuning, a process of extended training whereby the model is familiarized with a specific individual’s handwriting. By fine-tuning \textit{Transkribus} models, researchers are able to improve accuracy significantly, reporting CERs as low as 1.27\% and WERs as low as 5.97\%, although most fine-tuned results are in the range of CER 3-5\% and WER 12-20\% (Prebor 2023; Christlein et al 2018). But achieving these levels of accuracy on historical handwriting requires large amounts of training data. In one study, researchers had to manually transcribe 69,457 words (558 manuscript pages) to achieve a CER of 4.39\% and WER of 12.43\% (Ó Raghallaigh 2022). This may be feasible for projects involving a large corpus of single authored documents, as might be encountered by those working to digitize whole archival collections of personal or state papers (Sanchez et al 2019). At the same time, fine-tuning does not necessarily improve a model’s ability to generalize to more hands and can lead to “overfitting”, meaning that as a model gets better at reading a specific hand, it gets worse at other hands it has not seen (Perdiki 2023). Fine-tuning is thus task specific and usually must be repeated for each new hand in a given corpus. Yet this significantly limits real-world adoption of HTR by most historians who typically work on a variety of documents exhibiting a diversity of hands, styles, formatting, image types, and photographic attributes (Sánchez et al 2019). Many also lack the technological expertise or comfort-level necessary to engage with the fine-tuning process, even with the relatively user-friendly interfaces provided by programs like \textit{Transkribus}. 

The emergence of LLMs, such as OpenAI’s GPT-3 model in 2020, which exhibit remarkable capabilities in understanding and generating human-like text, potentially offers new possibilities in this direction as these models are built to be general purpose language processing machines (Brown 2020; Bommasani et al 2021). Their proficiency in zero-shot and few-shot learning, combined with unimaginably large corpuses of training data suggests they may be better able to generalize across a wider range of handwriting styles, formats, and languages than traditional HTR models. This raises the intriguing possibility that they might be harnessed to make HTR both more effective and accessible in real-world conditions.

\section{Large Language Models}

In the last few years, LLMs have significantly advanced the field of natural language processing, achieving remarkable performance across a variety of language tasks. Unlike traditional neural networks that are designed to learn narrowly defined tasks, LLMs are built upon the transformer architecture introduced by Vaswani et al. (2017) and are capable of generalizing to a wider distribution of data. This architecture employs self-attention mechanisms, allowing the model to weigh the importance of different words within a given context and generate coherent, contextually appropriate language, handling complexities like idioms, metaphors, and varying linguistic styles.

The LLM training process involves two main stages: pre-training and fine-tuning. During pre-training, the model is exposed to vast amounts of unlabelled text data, learning to predict the next token (equivalent to about 3/4s of a word) in a string—a process known as self-supervised learning (Radford et al., 2019; Brown et al., 2020). Training algorithms constantly adjust weighted parameters in the model’s neural network, using gradient descent to minimize error rates through a process called backpropagation (Rumelhart 1986). While it sounds complex, the process effectively allows the model to find settings (the weights) on its own that minimize the number of errors that it makes in predicting the next token. The larger the number of weighted parameters in a model, the more robust its statistical representation of language and its meaning (Kaplan 2020). While the process itself is relatively straightforward, it is impossible to comprehend the scale of the training data, the speed of the training process, or the complexity of the resulting statistical representation of language encoded in the model’s weights. Fine-tuning then adapts the behaviour of the pre-trained model to specific tasks using smaller, task-specific datasets (Ruder 2018). This contrasts with other types of neural networks like those traditionally applied to HTR that typically require large amounts of labelled data for each new task. 

While some historians have recently become interested in the potential application of LLMs to historical research, access to a corpus of accurately transcribed historical documents is essential (McLean et al 2024; Humphries and Leddy 2023). As transformer based LLMs were originally developed to aid in translation tasks—and have proved highly adept in that field (Vaswani et al 2017)—it seems likely that those same “skills” might transfer to transcription which shares many of the same features (Stevens and Burg 1997). Recent advancements have extended LLMs to handle multimodal inputs, integrating both textual and visual information. Models like OpenAI's GPT-4 (February 2023) are capable of processing and generating text based on visual inputs, including images of handwritten text (OpenAI 2023). This multimodal capability is particularly relevant for tasks like transcribing historical handwritten documents, where the model can leverage its language and contextual understanding to interpret and transcribe text from images without the need for extensive pre-processing, segmentation, or feature extraction (Humphries and Leddy 2024; Humphries and Story 2024). Despite this potential, the application of LLMs to handwriting recognition in historical documents remains unexplored (Humphries 2023). This study aims to fill this gap by examining how these models perform “out of the box” without fine-tuning on example documents typical of real-world use cases. 

\section{Transcription Pearl}

Our goal was to systematically test the ability of mainstream LLMs in the transcription of historical handwritten documents in real-world scenarios and to compare those transcriptions to a ground-truth document to determine error rates, comparing them to those reported for humans and HTR software. This required constructing a dataset that would simulate the typical working conditions of historians, archivists, and genealogists who often use a variety of documents written in diverse hands, captured in less-than-ideal conditions with phones or hand-held cameras as well as from black and white microfilm. 

To manage the workflow, we developed a software tool called \textit{Transcription Pearl}(\url{https://github.com/mhumphries2323/Transcription\_Pearl}) that allows users to easily and automatically process large numbers of images with an LLM through a Graphical User Interface (GUI) designed to visualize the process of transcribing and correcting a historical document (see Figure 1). Written in 3,907 lines of python code, users select a “plug and play” LLM of their choice from a settings menu to conduct the transcription. In our testing, these included all the major multi-modal frontier models, that is OpenAI’s gpt-4o-06-08-2024; Anthropic’s Claude Sonnet 3.5; and Google’s Gemini 1.5 Pro 002. While we would have liked to have tried open-source models like Llama-3, we did not have the requisite computing resources to run models like Llama 3.2 locally (McLean, 2024). As our goal was to test real-world use cases, the commercially available models are also the most accessible and economical models available. 

In \textit{Transcription Pearl}, users import locally stored images in jpg/jpeg formats through a drag and drop interface or extract them from PDFs. After they are imported, the images are automatically evaluated and, if necessary, resized so that no one side of the image is larger than 2048 pixels to conform with the restrictions on image size imposed by the LLMs. Behind the scenes, these images are stored in a \textit{Pandas} dataframe with corresponding fields to hold a “raw transcription” and “corrected transcript” for each image (along with other metadata). This is, in essence, a spreadsheet with an image of each page in the first field of each row followed by two blank fields where we store the transcription and its corrected version respectively. These transcription columns can later be collated and exported as text file documents.

The LLMs are responsible for generating the transcriptions and their corrections. The program “calls” the selected LLM via an Application Programming Interface (API), that is a secure electronic communications mechanism in which requests are sent to a server for processing, returning a response. In this case, the request contains a “prompt” giving the LLM its task along with the necessary information to complete that task and an image of the document. The LLM processes the request on an external server belonging to OpenAI, Anthropic, or Google and this server returns a response containing a number of messages in JSON (JavaScript Object Notation) format. Along with other data, the messages returned by the server include the LLM’s transcription (or corrected transcript) which is then extracted from the response. API errors, timeouts, and non-standard replies are handled with retry mechanisms, meaning the request is resent until a valid response is received or it exceeds three tries.

\begin{figure}
    \centering
    \begin{tcolorbox}[colback=white,boxrule=1pt]
        \includegraphics[width=0.98\linewidth]{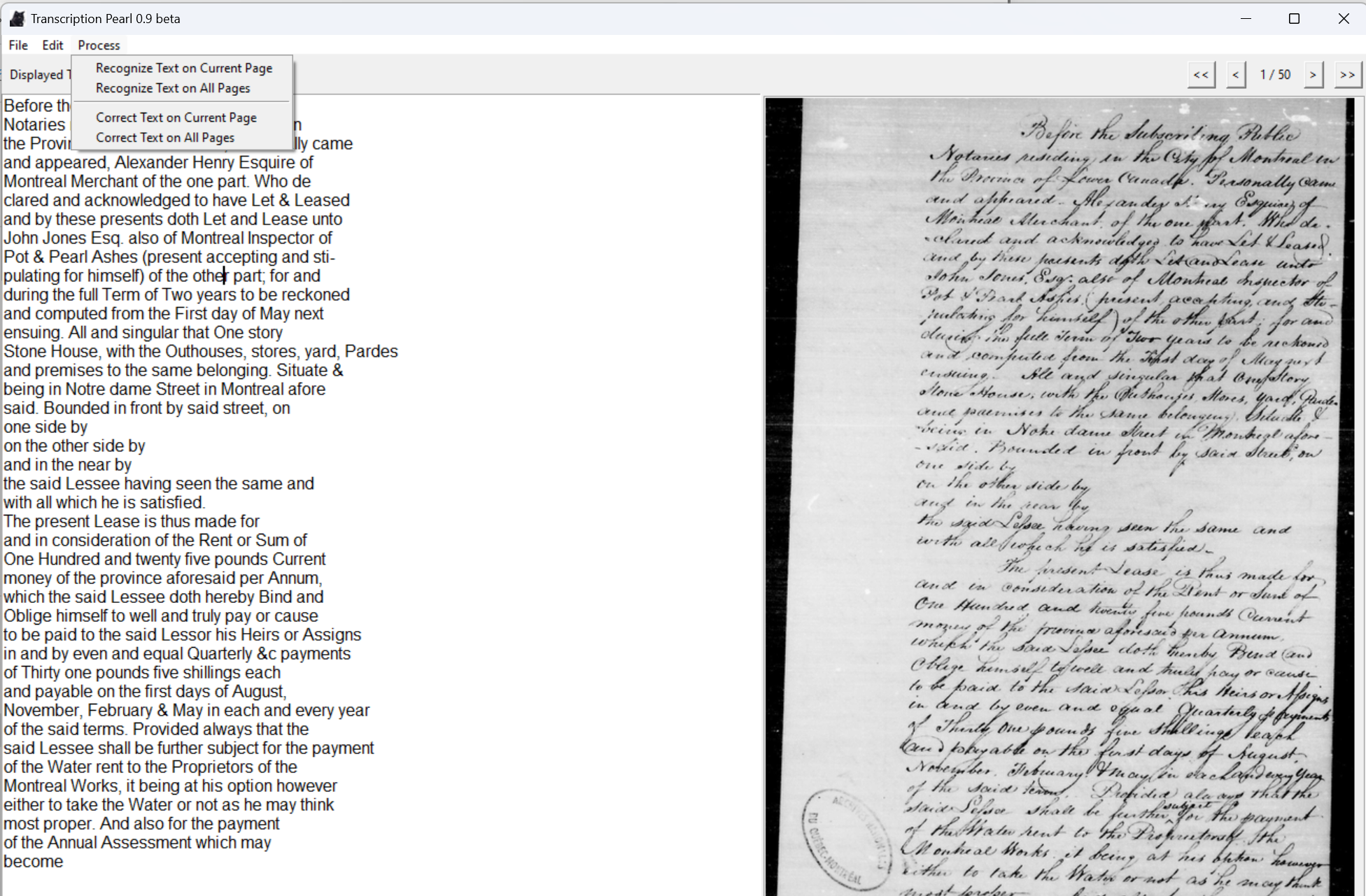}
    \end{tcolorbox}
    \caption{Figure 1: Transcription Pearl Interface}
    \label{fig:enter-label}
\end{figure}

To speed up the process, \textit{Transcription Pearl} makes the API calls asynchronously through parallel processing using the \textit{Asyncio} and \textit{ThreadPoolExecutor} libraries. This means that rather than waiting for each page to process sequentially, one at a time, we process all the pages at the same time. Progress is displayed to the user via a progress bar window while any error messages are conveyed via message boxes. Behind the scenes, the transcriptions extracted from the LLM’s response are collated and stored in the relevant column of the Pandas dataframe, displayed to the user in an editable textbox next to an image of the original handwritten text (see Figure 1). Users can click through each page in the document and can perform a variety of common word processing tasks on the text, including find and replace functions, eventually saving it and exporting it to a text file or PDF. 
\begin{figure}
    \centering
    \fbox{\includegraphics[width=0.98\linewidth]{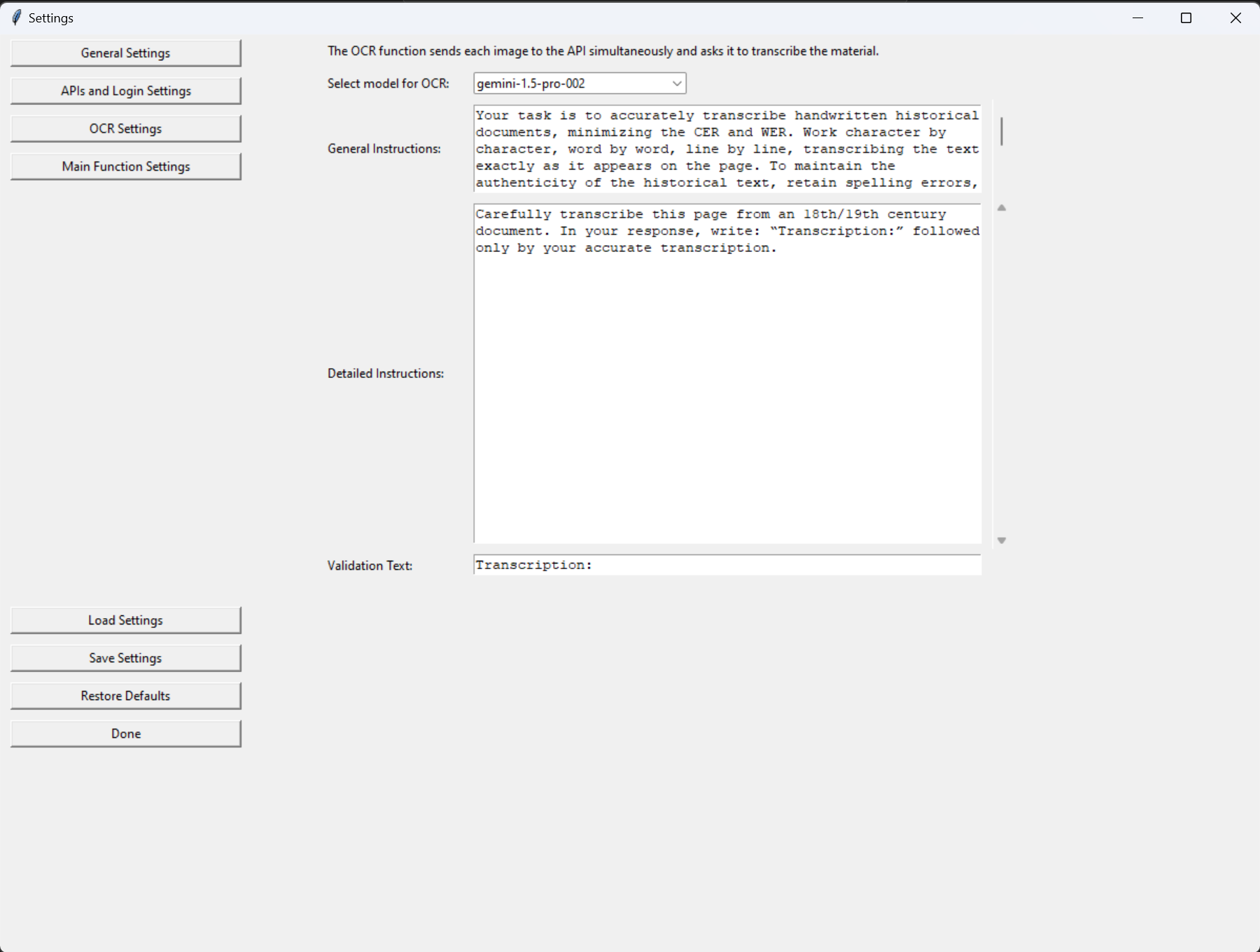}}
    \caption{Transcription Pearl Settings Window  }
    \label{fig:enter-label}
\end{figure}
It is important to understand that compared to chatbot interfaces like ChatGPT and Claude, the API interfaces to the LLMs provide significantly more control and customization over behaviour by allowing the programmer to adjust the prompt and various hyperparameters or settings. An LLM prompt is comprised of a customized System Message, which tells the model what it is supposed to do and how it is supposed to behave, and a User Message which gives the model the actual images and/or text of the documents, and any additional necessary context. Attention to prompting is essential and is increasingly being recognized as a way of programming via natural language. As Schulhoff et al have shown, there are many prompting strategies available, but the main ones include zero-shot techniques in which a model is simply told to perform a given task; in-context learning where the prompt includes examples of the task it is to perform (a form of fine-tuning on the fly); chain of thought in which models are told to “think” through problems and work “step by step”; as well as tree of thought where multiple possible solutions are explored and the best response is selected (Schulhoff et al 2024). All of these approaches can be highly sensitive to wording as well as the order of directions. 

To establish a baseline, in our tests we employed a simple zero-shot prompting technique. In \textit{Transcription Pearl}, users can experiment with different system and user messages in a settings menu or load prompts from stored settings files (see Figure 2). We expect that further experimentation could significantly improve our results, most likely through few-shot prompting focused on providing a small number of example pages of a given hand or format (Brown et al 2020). Further research in this direction is thus warranted.

In terms of hyperparameters, both temperature and Top-P are settings used to adjust the determinism or variability in the model’s outputs as well as the range of token choices available. Normally ranging from 0.0 to 1.0, temperature settings control how often the model chooses the most probable next token during “inference” (the generation of its response) with higher settings producing less probable outputs and lower settings making the model’s behaviour more deterministic. Top-P controls whether a model can choose from all the available tokens in its training data (1.0) or only the nth most common tokens (ie. a setting of 0.3 would constrain its choices to the top 30\% most common tokens). In all our tests aimed at establishing a baseline, we set the temperature to 0.0 to make the model outputs as consistent as possible, while leaving Top-P at the default setting of 1.0. Studies suggest that on non-creative tasks, lower temperature and standard Top-P settings perform best, although more research in this area is again warranted (Zhou et al 2024).

For the transcription function, our prompt was as follows:

\begin{quote}System Message: “Your task is to accurately transcribe handwritten historical documents, minimizing the CER and WER. Work character by character, word by word, line by line, transcribing the text exactly as it appears on the page. To maintain the authenticity of the historical text, retain spelling errors, grammar, syntax, and punctuation as well as line breaks. Transcribe all the text on the page including headers, footers, marginalia, insertions, page numbers, etc. If these are present, insert them where indicated by the author (as applicable). In your response, write: “Transcription:” followed only by your accurate transcription”

 User Message: “Carefully transcribe this page from an 18th/19th century document. In your response, write: “Transcription:” followed only by your accurate transcription.

 \{Image Inserted Here\}”
 \end{quote}

In the correction task, the models were given the following prompt:
 
\begin{quote}System Message: “Your task is to compare handwritten pages of text with corresponding draft transcriptions, correcting the transcription to produce an accurate, publishable transcript. Be sure that the spelling, syntax, punctuation, and line breaks in the transcription match those on the handwritten page to preserve the historical integrity of the document. Numbers also easily misread, so pay close attention to digits. You must also ensure that the transcription begins and ends in the same place as the handwritten document. Include any catchwords at the bottom of the page. In your response write “Corrected Transcript:” followed by your corrected transcription.”

 User Message: Your task is to use the handwritten page image to correct the following transcription, retaining the spelling, syntax, punctuation, line breaks, catchwords, etc of the original.

 In your response write “Corrected Transcript:” followed by your corrected transcription. 

 <Rough Transcription>\{Text of Rough Transcription Inserted Here\}</Rough Transcription>

 \{Image Inserted Here\}”
\end{quote}

\section{Testing Dataset}

To test the accuracy of the transcription and correction functions in real-world settings, we required access to diverse images of a range of types of historical handwritten documents with matching proofed transcriptions to constitute a “ground truth” document. While there are a number of standardized HTR datasets available such as the George Washington, READ 2016, IAM, or RIMES sets (AL Kendi, 2024), none of these simulate real-world conditions. More importantly, we also questioned whether the models may have been exposed to this data in training which would likely artificially deflate our reported error rates.

We found specific evidence of pre-exposure not only to standard training sets but also to most of the easily accessible collections of documents on the internet. This came from early tests we conducted with Google’s Gemini-1.5-Pro-001 model in which many of our API requests failed when the LLM refused to transcribe the documents, providing a “stop” reason of “recitation”, meaning the model stopped before providing its transcription. According to Google’s documentation, this safeguard was put in place to prevent the model from outputting materials that were in its training data (Google 2024). As a result, when users receive the “recitation” stop reason, it means that the Gemini model was trained on either the images of the documents, the accompanying transcriptions, or both. While false positives are certainly possible (as we discuss below), we found that nearly every document from well known collections at McGill University, Harvard University Archives, and the William Clements Library at the University of Michigan we tried were rejected. Although all the frontier LLMs are thought to have been trained on similar datasets, only Gemini notifies users when a request might expose that data. Therefore, in the absence of specific information about the contents of model training data, we must assume that if a document was on the internet prior to a model’s training cutoff, it likely appeared in training data.

It is essential to avoid including images or text in a testing dataset that also appeared in the model’s training data because it has the potential to artificially improve the accuracy of the model’s responses. In essence, if a model saw the combination of images and text in training, it was already trained on that specific handwriting. Even if it only saw the text, the model would be likely to begin to “recite” a transcription from memory rather than using its ability to “read” the actual handwriting. Both would lead to falsely low CERs and WERs. Indeed, in comparing the scores achieved on these early compromised datasets with those reported below, we found that the compromised datasets had CERs and WERs several standard deviations below those of the clean datasets we eventually developed.

As our intent was to capture the various types of documents that an active historian might be expected to encounter in a typical research project, we thus created our own heterogeneous corpus of fifty pages (9,911 words) of letters, memoranda, diary entries, and legal documents written in English between 1761 and 1827 and featuring 33 different hands. We selected a diverse range of document types and formats typically encountered by historians: 13 are scans from black and white microfilms, 26 are medium to high quality scans/photographs, and 11 are photographs taken with an iPhone or similar handheld device. All are 96 dpi with resolutions between 1742x2048 and 1222x2048. Six of the documents are somewhat blurry, while eight were shot in a darkly lit room, producing an image with poor contrast between the text and page. The hands featured in the documents range from elegant formal writing to the highly vernacular. In the former, capitalization, punctuation, and spelling were somewhat standardized (by the standards of the day at least) while in the latter they were often highly variable. The ground-truth test-set was developed from images that members of the team had acquired during previous research projects and so pertained mainly to the North American fur trade. To the best of our knowledge, they have never been published on the internet, are not publicly available, and so could not have been included in training data. While our dataset was designed to simulate real-world conditions, we recognize that it is relatively small and is focused solely on English language documents from the 18th and 19th centuries. 

Even so, one of our documents did occasionally raise a false positive with the Gemini-1.5-Pro-002 (but not 001) model, which cited a specific website in its response where the text was supposed to have appeared (\url{https://sites.google.com/site/longpointsettlers/norfolk-history/john-dease}). Initially this elicited some confusion as a thorough search of that website revealed that it reproduced none of our documents. However, we realized that both the website and our dataset included 19\textsuperscript{th} century English notarial records written in Montreal around the same time which contained some superficially similar wording in the \textit{pro forma} sections of the documents (ie. “Before the subscribing Public Notaries residing in the City of Montreal in the province of Lower Canada, Personally Appeared…”). The similar structures and appearance of the documents appears to have occasionally caused this document to be flagged for recitation, but it was an inconsistent finding, roughly 1 in 30 times.

We created a ground truth document using an automated transcription generated with \textit{Transkribus} and manually corrected and proofed. Marginalia was inserted in the text where indicated by the author and struck-through text was excluded. Dates, salutations, abbreviations, and page numbers were included where they appeared in the document.

In terms of data privacy and security, all the documents we used in our testing dataset were out of copyright and in the public domain. The terms of service for the APIs we used in our testing also precludes OpenAI, Anthropic, and Google from retaining or using this data to train their models in future. Those APIs are also all certified to comply with most protocols and regulations governing data privacy and security. We have decided not to make the dataset public to preserve its integrity for testing the efficacy of future models. If it were uploaded to the internet, it is highly likely to appear in the training datasets of future models and would no longer be useful for testing purposes. We expect that a similar corpus of English language documents should elicit similar results. 

\section{Testing and Evaluation Metrics}

For testing purposes, WER and CER error rates were calculated using a python script (see \url{https://github.com/mhumphries2323/Transcription\_Pearl/settings}) employing the python \textit{Jiwer} , D\textit{ifflib, NLTK, Inflect, }and \textit{Enchant} libraries to compare a proofed ground truth text with a hypothesis text automatically. The WER was calculated using the standard formula:  where S = number of substitutions (words incorrectly transcribed); D = number of deletions (words omitted from the transcription); I = number of insertions (extra words added to the transcription); and N = total number of words in the reference text. Accuracy was verified by adding the total number of correct words to the sum of S, D, and I and subtracting the result from N to obtain 0 if the analysis was correct. Similarly, the CER was computed as:  where S, D, and I are defined as above, but at the character level; T = total number of characters in the reference text. Results are validated in a similar way as with WER, albeit by adding correct characters to the sum of S, D, I and subtracting T. At the end of the script, the list of error words and the corresponding correct or “ground truth” word, and number of instances of each error, were outputted to a CSV to facilitate further qualitative analysis.

To evaluate the accuracy of the transcriptions both for testing purposes and for real-world applications, we used two different CER/WER metrics employing different definitions of S (that is, of what constituted a “substitution”): a strict CER/WER and modified CER/WER. With the strict measures, any difference in capitalization, punctuation, or spelling between the original document and transcription was counted as an error for both CER and WER. With the modified measure of CER and WER, we ignored changes in capitalization and punctuation and did not count substitutions as errors where a word was spelled incorrectly in the ground truth document and then corrected in the transcription (i.e. “employ’d” in the original became “employed” in the transcript). Any other type of substitutions (including any differences in digits), insertions or deletions were counted as errors in both measures. The script allows the user to switch between both modes for the same document.

Using two different measures of error allowed us to compare LLM performance against standard metrics in the field (the strict CER/WER) as well as to evaluate their usefulness in real-world applications where strict adherence to historical capitalization, punctuation, and spelling may not be required. For both metrics, we standardized the texts by eliminating excess whitespace (ie extra spaces, empty lines, etc.) from the ground-truth and hypothesis documents. The list of errors for each comparison was then compiled in a CSV spreadsheet and saved for further qualitative analysis. 

For each test, we ran the set of fifty documents through \textit{Transkribus}’s newest and best “Titan Super Model” as well as the relevant function(s) in the \textit{Transcription Pearl} program, ten times each for a total of 500 pages per model, per test (IE 50 page test set x 10 tests = 500 pages). We then averaged the results across all ten outputs per model per test, reported high and low scores, and calculated standard deviations for each run. We also recorded latency (time to completion) as well as cost.

\section{Performance on Out of the Box Transcription Tasks}

In our testing, we found that frontier model LLMs produce more accurate transcriptions of handwritten historical English language documents than state-of-the-art HTR models on “out-of-the-box” transcription tasks (Tables 1 and 2). To establish a baseline for conventional HTR software, we looked at both transcriptions generated with Trankribus’s older PyLaia model (which is the model most featured in the literature) and its newer and more advanced transformer based Titan Super Model. The PyLaia model transcribed the test data set with an average strict CER of 10.3\% and WER of 27.0\% which is comparable to the results reported by other authors (Christlein et al 2018; Ó Raghallaigh 2022; Prebor 2023). The Titan model improved on these results by 20\% (CER=8.0\%) and 26\% (WER=19.7\%). This means that on the task of achieving a perfect transcription, the \textit{Transkribus} models correctly transcribed between 90 and 92\% of the characters and 73 and 80\% of the words in the ground truth document. 

In terms of the LLMs, two of the three models tested outperformed the PyLaia model on strict CER while all three did better on strict WER. Only Claude Sonnet-3.5 improved on the Transkibus Titan Super Model, scoring 10\% better on character accuracy and 19\% on word accuracy, achieving a strict CER of 7.3\% and strict WER of 15.9\% (Gemini 1.5-Pro-002 also outperformed Titan by 4\%). This shows that frontier LLMs can achieve state-of-the-art performance without fine-tuning or training on specific document formats or handwriting styles. While this was significantly better than the conventional HTR results, we should note that it is still about 60\% less accurate than the upper error rates reported for non-expert human transcribers (Feng et al 2020; Nordo et al 2017; Oliveira 2018; Stolcke 2017).

\begin{table}[ht]
\centering
\resizebox{\textwidth}{!}{
\begin{tabular}{|l|c|c|c|c|}
\hline
Model & Average CER \% (n=10) & Std. Dev. & High (\%) & Low (\%) \\
\hline
Transkribus (PyLaia) & 0.103 & 0.0 & 0.103 & 0.103 \\
Transkribus (Titan) & 0.080 & 0.001 & 0.080 & 0.075 \\
Claude Sonnet-3.5 & 0.073 & 0.001 & 0.074 & 0.072 \\
gpt-4o-06-08-2024 & 0.110 & 0.001 & 0.113 & 0.108 \\
Gemini-1.5 Pro-002 & 0.085 & 0.003 & 0.089 & 0.079 \\
\hline
\end{tabular}
}
\caption{Strict Character Error Rates (CER) of Transcriptions by Model}
\end{table}
\begin{table}[ht]
\centering
\resizebox{\textwidth}{!}{
\begin{tabular}{|l|c|c|c|c|}
\hline
Model & Average WER \% (n=10) & Std. Dev. & High (\%) & Low (\%) \\
\hline
Transkribus (PyLaia) & 0.270 & 0.0 & 0.270 & 0.270 \\
Transkribus (Titan) & 0.197 & 0.001 & 0.197 & 0.193 \\
Claude Sonnet-3.5 & 0.159 & 0.001 & 0.161 & 0.158 \\
gpt-4o-06-08-2024 & 0.230 & 0.002 & 0.234 & 0.228 \\
Gemini-1.5 Pro-002 & 0.188 & 0.004 & 0.194 & 0.182 \\
\hline
\end{tabular}
}
\caption{Strict Word Error Rates (WER) of Transcriptions by Model}
\end{table}

As indicated in Tables 3 and 4, when differences in punctuation, capitalization, and the correction of historical spelling errors are excluded from the error metrics, the performance of all the models improved significantly. For PyLaia the modified CER was 23\% lower than the stricter measure while modified WER was 38\% less (7.6 and 17.4\%). For Titan, those measures fell 18 and 33\% to 6.6 and 13.2\% respectively. Another way of interpreting these results is to say that roughly 38\% (PyLaia) and 33\% (Titan) of the initial word-level errors were actually differences in capitalization, punctuation, or historical versus modern spelling. 

The results were similar with the LLMs we tested, with modified CERs falling 19-22\% and modified WER decreasing by 39-44\%. Claude Sonnet-3.5 was the most accurate, achieving a modified CER of 5.7\% and modified WER of 8.9\% meaning it got more than 91\% of the actual words correct, even if it standardized punctuation, capitalization, and spelling. On the whole, though, this suggests that LLMs correctly transcribed roughly 20\% more words than conventional HTR models.
\begin{table}[ht]
\centering
\resizebox{\textwidth}{!}{
\begin{tabular}{|l|c|c|c|c|}
\hline
Model & Average CER \% (n=10) & Std. Dev. & High (\%) & Low (\%) \\
\hline
Transkribus (PyLaia) & 0.076 & 0.0 & 0.076 & 0.076 \\
Transkribus (Titan) & 0.066 & 0.001 & 0.066 & 0.062 \\
Claude Sonnet-3.5 & 0.057 & 0.001 & 0.058 & 0.056 \\
gpt-4o-06-08-2024 & 0.090 & 0.001 & 0.092 & 0.087 \\
Gemini-1.5 Pro-002 & 0.067 & 0.003 & 0.071 & 0.060 \\
\hline
\end{tabular}
}
\caption{Modified Character Error Rates (CER) on Transcriptions by Model}
\end{table}

\begin{table}[ht]
\centering
\resizebox{\textwidth}{!}{
\begin{tabular}{|l|c|c|c|c|}
\hline
Model & Average WER \% (n=10) & Std. Dev. & High (\%) & Low (\%) \\
\hline
Transkribus (PyLaia) & 0.174 & 0.0 & 0.174 & 0.174 \\
Transkribus (Titan) & 0.132 & 0.001 & 0.132 & 0.130 \\
Claude Sonnet-3.5 & 0.089 & 0.001 & 0.091 & 0.087 \\
gpt-4o-06-08-2024 & 0.141 & 0.002 & 0.144 & 0.136 \\
Gemini-1.5 Pro-002 & 0.110 & 0.004 & 0.117 & 0.102 \\
\hline
\end{tabular}
}
\caption{Modified Word Error Rates (WER) on Transcriptions by Model}
\end{table}
In terms of latency, as outlined in Table 5, \textit{Transkribus} took about 22 minutes to transcribe all 50 pages, or around 40 seconds per page. In contrast, all three LLMs completed a batch of 50 transcriptions in around 30 seconds or roughly 50 times faster than \textit{Transkribus}. Regarding cost, automated LLM transcriptions cost around 1,500 times less than human transcriptions (albeit with an error rate that is 10 times higher than those advertised by many professional human services) and roughly 50 times less (between 29 and 70 times, depending on the model) than with comparable, automated HTR services.
\begin{table}[ht]
\centering
\resizebox{\textwidth}{!}{
\begin{tabular}{|l|c|c|c|c|c|c|c|}
\hline
Model & Input & Output & Cost per Million & Total \$ / & Average Time to Completion & Std. Dev. & Seconds \\
 & Tokens & Tokens & Input/Output Tokens & Per Page & (50 Pages) / Seconds Per Page (spp) & (seconds) & Per Page \\
 & & & or Per Page & & & & \\
\hline
Transkribus (PyLaia) & -- & -- & \$0.26 per page & \$13.00 /\$0.26 & -- & -- & -- \\
Transkribus (Titan) & -- & -- & \$0.26 per page & \$13.00 /\$0.26 & 1179.4 seconds (19.65 Minutes)/ 23.58 spp & 543.95 & 23.58 \\
Claude Sonnet-3.5 & 87,000 & 15,000 & \$3.00 / \$15.00 & \$0.48/\$0.0096 & 35.74 seconds / 0.71 spp & 0.76 & 0.71 \\
gpt-4o-06-08-2024 & 57,000 & 13,500 & \$2.50 / \$10.00 & \$0.28/\$0.0052 & 28.60 seconds / 0.57 spp & 11.66 & 0.57 \\
Gemini-1.5 Pro-002 & 21,200 & 13,500 & \$3.50 / \$10.50 & \$0.19/\$0.0038 & 28.08 seconds / 0.56 spp & 3.2 & 0.56 \\
\hline
\end{tabular}
}
\caption{Latency (Time to Completion) of Transcription and Cost in USD}
\end{table}

\section{LLM Transcription Correction }

Unlike conventional HTR models that can transcribe text but are unable to correct, it, we also found that frontier model LLMs could also be employed to significantly improve error rates by comparing images of the original handwritten pages to the text of LLM generated transcriptions to produce new, corrected transcripts. Interestingly, this capability was limited to heterogeneous model interactions, that is different transcription and correction models. In other words, LLMs showed no statistically significant ability to self-correct their own outputs which is consistent with other work on LLM error correction (Huang 2024).

As shown in Tables 6 and 7, on average the corrected transcripts created by the three LLMs we tested showed improvements in strict CER of 13\% and WER of 10\%. Results varied widely depending on the combination of models employed. Claude Sonnet-3.5 recorded the best overall scores when correcting Gemini transcriptions, achieving an average strict CER of 5.7\% and WER of 13.8\%. Although Claude Sonnet-3.5 also achieved the best overall reduction (25.5\%) in the word error rates of Gemini 1.5-Pro-002 generated transcriptions, it only managed to reduce errors in the gpt-4o-06-08-2024 transcriptions by 17.4\% (see discussion below).

\begin{sidewaystable}
\centering
\small
\begin{tabular}{|l|l|c|c|c|c|c|}
\hline
Initial Transcription Model & Correction Model & Average CER \% (n=10) & Std. Dev. & High (\%) & Low (\%) & Improvement from Initial CER (\%) \\
\hline
Claude Sonnet-3.5 & Gemini-1.5 Pro-002 & 0.057 & 0.001 & 0.059 & 0.055 & 0.329 \\
Gemini-1.5 Pro-002 & Claude Sonnet-3.5 & 0.058 & 0.001 & 0.060 & 0.056 & 0.206 \\
gpt-4o-06-08-2024 & Claude Sonnet-3.5 & 0.067 & 0.001 & 0.068 & 0.066 & 0.081 \\
gpt-4o-06-08-2024 & Gemini-1.5 Pro-002 & 0.074 & 0.002 & 0.076 & 0.070 & 0.132 \\
Claude Sonnet-3.5 & Claude Sonnet-3.5 & 0.071 & 0.001 & 0.072 & 0.070 & 0.023 \\
Gemini-1.5 Pro-002 & gpt-4o-06-08-2024 & 0.079 & 0.007 & 0.098 & 0.074 & 0.282 \\
Gemini-1.5 Pro-002 & Gemini-1.5 Pro-002 & 0.084 & 0.003 & 0.090 & 0.080 & 0.012 \\
Claude Sonnet-3.5 & gpt-4o-06-08-2024 & 0.098 & 0.016 & 0.136 & 0.084 & 0.112 \\
gpt-4o-06-08-2024 & gpt-4o-06-08-2024 & 0.112 & 0.005 & 0.127 & 0.108 & -0.014 \\
\hline
\end{tabular}
\caption{Strict Character Error Rates (CER) on LLM to LLM Corrected Transcriptions by Model}
\vspace{1cm}

\begin{tabular}{|l|l|c|c|c|c|c|}
\hline
Initial Transcription Model & Correction Model & Average WER \% (n=10) & Std. Dev. & High (\%) & Low (\%) & Change from Initial WER (\%) \\
\hline
Gemini-1.5 Pro-002 & Claude Sonnet-3.5 & 0.138 & 0.002 & 0.141 & 0.136 & 0.131 \\
Claude Sonnet-3.5 & Gemini-1.5 Pro-002 & 0.140 & 0.002 & 0.142 & 0.137 & 0.255 \\
gpt-4o-06-08-2024 & Claude Sonnet-3.5 & 0.153 & 0.001 & 0.155 & 0.151 & 0.042 \\
Claude Sonnet-3.5 & Claude Sonnet-3.5 & 0.156 & 0.001 & 0.158 & 0.155 & 0.018 \\
gpt-4o-06-08-2024 & Gemini-1.5 Pro-002 & 0.171 & 0.002 & 0.174 & 0.166 & 0.094 \\
Claude Sonnet-3.5 & gpt-4o-06-08-2024 & 0.190 & 0.016 & 0.229 & 0.174 & 0.174 \\
Gemini-1.5 Pro-002 & gpt-4o-06-08-2024 & 0.179 & 0.007 & 0.199 & 0.175 & 0.225 \\
Gemini-1.5 Pro-002 & Gemini-1.5 Pro-002 & 0.185 & 0.003 & 0.190 & 0.179 & 0.018 \\
gpt-4o-06-08-2024 & gpt-4o-06-08-2024 & 0.232 & 0.005 & 0.245 & 0.228 & -0.009 \\
\hline
\end{tabular}

\caption{Strict Word Error Rates (WER) on LLM to LLM Corrected Transcriptions by Model}
\end{sidewaystable}

When scores are modified to exclude differences in capitalization and punctuation as well as historical spelling corrections (see Tables 8 and 9), Claude Sonnet-3.5 again achieved the best overall performance, correcting Gemini transcriptions to a CER of 4.1\% and WER of 7.0\%. This represented reductions of the initial modified error rates of 63\% and 38\% respectively. These represent truly significant improvements that show LLMs are capable of producing transcriptions that are more than 96\% accurate at the character level and 93\% at the word level, scores that are in the realm of non-expert human transcribers. Again, though, model performance varied depending on the combination of models employed. On average, they improved the modified CER by 36\% and the WER by 16\%.

We also found that models were incapable of homogenous error correction. Although not statistically significant, gpt-4o-06-08-2024 scores actually declined by between 1 and 2\% on both strict and modified metrics when it was asked to correct its own transcriptions. On average, all model scores only changed about 1\% on self-correction. This is in stark contrast to the 15-20\% improvement on both strict and modified CER and 22-23\% improvement on WER for heterogeneous corrections. 
\begin{sidewaystable}
\centering
\small
\begin{tabular}{|l|l|c|c|c|c|c|}
\hline
Initial Transcription Model & Correction Model & Average CER \% (n=10) & Std. Dev. & High (\%) & Low (\%) & Improvement from Initial CER (\%) \\
\hline
Claude Sonnet-3.5& Gemini-1.5 Pro-002& 0.041& 0.002& 0.044& 0.039& 0.387\\
Gemini-1.5 Pro-002& Claude Sonnet-3.5& 0.042& 0.001& 0.044& 0.040& 0.261\\
gpt-4o-06-08-2024& Claude Sonnet-3.5& 0.050& 0.001& 0.052& 0.049& 0.111\\
gpt-4o-06-08-2024& Gemini-1.5 Pro-002& 0.056& 0.002& 0.058& 0.053& 0.170\\
Claude Sonnet-3.5& Claude Sonnet-3.5& 0.055& 0.001& 0.056& 0.054& 0.025\\
Gemini-1.5 Pro-002& gpt-4o-06-08-2024& 0.060& 0.007& 0.079& 0.055& 0.330\\
Gemini-1.5 Pro-002& Gemini-1.5 Pro-002& 0.066& 0.003& 0.072& 0.062& 0.018\\
Claude Sonnet-3.5& gpt-4o-06-08-2024& 0.081& 0.016& 0.118& 0.067& 0.096\\
gpt-4o-06-08-2024& gpt-4o-06-08-2024& 0.091& 0.006& 0.107& 0.087& -0.015\\
\hline
\end{tabular}
\caption{Modified Character Error Rates (CER) on LLM to LLM Corrected Transcriptions by Model}

\vspace{1cm}

\begin{tabular}{|l|l|c|c|c|c|c|}
\hline
Initial Transcription Model & Correction Model & Average WER \% (n=10) & Std. Dev. & High (\%) & Low (\%) & Change from Initial WER (\%) \\
\hline
Claude Sonnet-3.5& Gemini-1.5 Pro-002& 0.070& 0.002& 0.073& 0.065& 0.367\\
Gemini-1.5 Pro-002& Claude Sonnet-3.5& 0.069& 0.001& 0.070& 0.067& 0.229\\
gpt-4o-06-08-2024& Claude Sonnet-3.5& 0.080& 0.002& 0.083& 0.079& 0.098\\
gpt-4o-06-08-2024& Gemini-1.5 Pro-002& 0.090& 0.002& 0.094& 0.085& 0.180\\
Claude Sonnet-3.5& Claude Sonnet-3.5& 0.087& 0.001& 0.088& 0.086& 0.022\\
Gemini-1.5 Pro-002& gpt-4o-06-08-2024& 0.094& 0.007& 0.115& 0.089& 0.330\\
Claude Sonnet-3.5& gpt-4o-06-08-2024& 0.117& 0.017& 0.155& 0.100& 0.171\\
Gemini-1.5 Pro-002& Gemini-1.5 Pro-002& 0.106& 0.003& 0.109& 0.101& 0.039\\
gpt-4o-06-08-2024& gpt-4o-06-08-2024& 0.142& 0.006& 0.160& 0.138& -0.011\\
\hline
\end{tabular}
\caption{Modified Word Error Rates (WER) on LLM to LLM Corrected Transcriptions by Model}
\end{sidewaystable}

Research on LLM self-correction is a highly technical and evolving field (Pan 2024). While little has been written on heterogenous vs homogeneous correction, a models’ failure to self-correct may be explained by the fact that when tasked with correcting a text, an LLM does not search for errors in the same way a human editor would (Liang 2024; Stechly 2023). Instead, at a basic level, they produce predictions about the most probable possible text outputs given an instruction (“correct the errors in the following document”), a text, and an image. It is plausible that this would limit the model’s ability to correct its own transcription errors because it must rely on the same set of weights that produced the mistakes in the first place in order to find errors, preventing the model from recognizing and rectifying them. In other words, the errors persist because they are not actually anomalies in terms of the model’s underlying statistical matrices: they indeed represent a probable output, just not the correct one. This suggests that models might be inherently biased toward their own outputs, making them less effective at self-correction compared to correcting texts produced by different models based on a different set of weights. An important area for further research is whether this tendency in transcription transfers to other areas of model behaviour and whether models might be intentionally trained or fine-tuned in a way that would allow them to overcome these biases and improve self correction.

Interestingly, the predictive nature of LLMs can also be harnessed to achieve human-level results when they correct transcriptions generated by traditional HTR systems like \textit{Transkribus}. Because \textit{Transkribus} outputs via the PyLaia and Titan models are based on a combination of HTR techniques that include traditional character level recognition as well as processing via transformer based neural networks (in the case of Titan), the transcriptions have an entirely different set of statistical and linguistic properties than those generated by LLMs. When these transcriptions are processed by an LLM for correction (along with the handwritten image), they activate a different set of probabilities than would be the case with homogeneous inputs—and these are likely different too from those of other LLMs. For example, in its own transcriptions and corrections, an LLM might sometimes choose the word “furs” over the historical spelling “furrs” because the former was a more probable output learned from its training data. However, if a \textit{Transkribus} transcription contained a misread character, as in the case of “furr5”, the most probable output might actually be “furrs”. An LLM is unlikely to make the first error, based on its training data, and so can more easily recognize how to correct it. Given that heterogeneous model combinations tended to produce better results in our earlier tests, it stands to reason that the error rates achieved with LLM corrections of \textit{Transkribus} transcriptions might also be significantly lower. 

\section{Using LLMs to Correct HRT Transcriptions}

As shown in Table 10, employing LLMs such as Claude Sonnet-3.5 to correct \textit{Transkribus} outputs led to a significant 59\% reduction in strict WER, achieving a near-human level average WER of 9\%. The strict CER also decreased substantially by 61\%, resulting in an average CER of around 4\%. Notably, the modified WER shows the most dramatic improvements with Claude Sonnet-3.5 reducing character level errors in Titan transcriptions by 68\% to 1.8\% and word level errors in the initial \textit{Transkribus} PyLaia transcription by a remarkable 74\% to 3.5\%. Given the discussion above, it is notable that the LLMs performed best correcting the transcriptions generated by the older PyLaia models which employ traditional, non-transformer based approaches to HRT (in contrast with Titan). This makes the PyLaia model more likely to generate errors at the character level, similar to those you might expect to see with print-based OCR (i.e. “furr5” or “5everal”). As noted above, LLMs have an easier time detecting and correcting those types of errors than ones that arise from transformer-based models.
\begin{sidewaystable}
\centering
\resizebox{\textwidth}{!}{
\begin{tabular}{|l|l|c|c|c|c|c|}
\hline
\multicolumn{7}{|c|}{\textbf{Strict CER}} \\
\hline
Transcription Model & Correction Model & Average \% (n=10) & Std. Dev. & High (\%) & Low (\%) & Change from Initial CER/WER (\%) \\
\hline
Transkribus (Titan) & Claude Sonnet-3.5 & 0.030 & 0.000 & 0.031 & 0.030 & 0.60 \\
Transkribus (Titan) & Gemini-1.5 Pro-002 & 0.041 & 0.001 & 0.042 & 0.039 & 0.46 \\
Transkribus (PyLaia) & Claude Sonnet-3.5 & 0.040 & 0.000 & 0.041 & 0.040 & 0.61 \\
Transkribus (Titan) & gpt-4o-06-08-2024 & 0.042 & 0.001 & 0.043 & 0.041 & 0.45 \\
Transkribus (PyLaia) & Gemini-1.5 Pro-002 & 0.054 & 0.001 & 0.057 & 0.052 & 0.48 \\
Transkribus (PyLaia) & gpt-4o-06-08-2024 & 0.059 & 0.001 & 0.061 & 0.058 & 0.43 \\
\hline
\multicolumn{7}{|c|}{\textbf{Strict WER}} \\
\hline
Transkribus (Titan) & Claude Sonnet-3.5 & 0.090 & 0.000 & 0.091 & 0.089 & 0.54 \\
Transkribus (PyLaia) & Claude Sonnet-3.5 & 0.112 & 0.000 & 0.112 & 0.111 & 0.59 \\
Transkribus (Titan) & gpt-4o-06-08-2024 & 0.115 & 0.002 & 0.118 & 0.112 & 0.39 \\
Transkribus (Titan) & Gemini-1.5 Pro-002 & 0.120 & 0.001 & 0.122 & 0.117 & 0.42 \\
Transkribus (PyLaia) & gpt-4o-06-08-2024 & 0.144 & 0.002 & 0.150 & 0.141 & 0.40 \\
Transkribus (PyLaia) & Gemini-1.5 Pro-002 & 0.162 & 0.002 & 0.165 & 0.160 & 0.47 \\
\hline
\multicolumn{7}{|c|}{\textbf{Modified CER}} \\
\hline
Transkribus (Titan) & Claude Sonnet-3.5 & 0.018 & 0.000 & 0.019 & 0.018 & 0.68 \\
Transkribus (Titan) & Gemini-1.5 Pro-002 & 0.026 & 0.001 & 0.027 & 0.024 & 0.55 \\
Transkribus (PyLaia) & Claude Sonnet-3.5 & 0.026 & 0.000 & 0.026 & 0.026 & 0.66 \\
Transkribus (Titan) & gpt-4o-06-08-2024 & 0.027 & 0.001 & 0.028 & 0.026 & 0.54 \\
Transkribus (PyLaia) & Gemini-1.5 Pro-002 & 0.036 & 0.001 & 0.039 & 0.035 & 0.52 \\
Transkribus (PyLaia) & gpt-4o-06-08-2024 & 0.038 & 0.001 & 0.041 & 0.037 & 0.50 \\
\hline
\multicolumn{7}{|c|}{\textbf{Modified WER}} \\
\hline
Transkribus (Titan) & Claude Sonnet-3.5 & 0.035 & 0.000 & 0.036 & 0.034 & 0.72 \\
Transkribus (PyLaia) & Claude Sonnet-3.5 & 0.045 & 0.000 & 0.045 & 0.044 & 0.74 \\
Transkribus (Titan) & Gemini-1.5 Pro-002 & 0.047 & 0.002 & 0.050 & 0.044 & 0.62 \\
Transkribus (Titan) & gpt-4o-06-08-2024 & 0.047 & 0.001 & 0.049 & 0.046 & 0.63 \\
Transkribus (PyLaia) & Gemini-1.5 Pro-002 & 0.065 & 0.002 & 0.069 & 0.063 & 0.63 \\
Transkribus (PyLaia) & gpt-4o-06-08-2024 & 0.067 & 0.001 & 0.070 & 0.066 & 0.61 \\
\hline
\end{tabular}
}
\caption{Error Rates of Conventional HTR Generated Transcriptions Corrected by an LLM Model}
\end{sidewaystable}
With character level accuracy approaching 99\% and word-level accuracy above 96\%, these results would be immediately usable in most applications such as keyword or semantic search, to enhance accessibility, and improve readability. This also suggests that because nearly 4/5 of the errors remaining after correction were limited to punctuation, capitalization, and the correction of historical spelling, fine-tuning might significantly improve these results.
\begin{table}[ht]
\centering
\resizebox{\textwidth}{!}{
\begin{tabular}{|l|c|c|c|c|c|c|c|}
\hline
Model & Input & Output & Cost per Million & Total \$ / & Avg. Time to Completion & Avg. Std. Dev. & Avg. Seconds \\
 & Tokens & Tokens & Input/Output Tokens & Per Page\$ & (50 Pages) & (seconds) & Per Page \\
\hline
Claude Sonnet-3.5 & 102,000 & 15,000 & \$3.00 / \$15.00 & \$0.53 / \$0.01 & 76.17 seconds & 8.55 & 1.52 \\
gpt-4o-06-08-2024 & 70,500 & 13,500 & \$2.50 / \$10.00 & \$0.31/ \$0.0062 & 61.18 seconds & 20.08 & 1.22 \\
Gemini-1.5 Pro-002 & 34,700 & 13,500 & \$3.50 / \$10.50 & \$0.26/ \$0.0052 & 32.27 seconds & 5.2 & 0.65 \\
\hline
\end{tabular}
}
\caption{Latency (Time to Completion) for Corrected Transcriptions and Cost in USD}
\end{table}
In terms of cost, as indicated by Table 11, correcting 50 pages of text was slightly more expensive than completing an initial transcription because the original textual transcription was sent to the model in addition to the page image. Even so, those costs ranged from \$0.005 per page for Gemini-1.5-Pro-002 to \$0.01 per page for Claude Sonnet-3.5. Latency was also roughly doubled for Claude Sonnet-3.5 and gpt-4o-06-08-2024, but comparable for Gemini-1.5-Pro-002. In practice, this means that the approximate cost of generating a corrected transcript using the best model combinations (for correction and Transkribus Titan for transcription and Claude Sonnet-3.5 for correction) would cost \$0.0138 per page for a Gemini-1.5-Pro-002 transcription corrected by Claude Sonnet-3.5 and \$0.27 per page for a Transkribus Titan transcription corrected by Claude Sonnet-3.5. 

\section{Conclusion}

This study demonstrates that when LLMs are integrated into a software application like \textit{Transcription Pearl}, they can achieve state-of-the-art performance on historical handwritten English language documents without fine-tuning or extensive pre-processing. Tests on a diverse corpus of 18th and 19th century texts shows they consistently generate transcriptions with character level accuracy of between 93 and 94\% and word level accuracy of between 86 and 91\%, depending on how errors were counted, significantly outperforming specialized HRT software. Most significantly, though, when the workflow in the software program is extended to include a secondary correction task, LLMs can achieve non-expert human levels of accuracy: CERs of between 4 and 6\% and WERs of 7 and 14\%, depending on how differences in capitalization, punctuation, and historical spelling are counted. When LLMs are tasked with correcting conventional HRT transcriptions, they begin to approach the accuracy rates reported for experienced human transcribers: character accuracy between 97 and 99\% and word accuracy between 91 and 96\%. The fact that LLMs completed transcription and correction tasks 50 times faster and at roughly 1/50th the cost of commercially available HTR software also makes them a highly accessible and cost-effective solution for mass digitization projects as well as for individual researchers.

The ability of LLMs to achieve these results without fine-tuning or training on specific hands or document formats is especially notable. It suggests LLMs are more flexible and adept at handling the variations in handwriting, formatting, and image quality typical of real-world historical documents compared with traditional HTR models that often struggle to generalize to new hands or document types without extensive fine-tuning. While further testing on documents from other time periods and languages may show different results, this study provides strong evidence that LLMs represent a promising new approach to the transcription of historical handwritten documents. By achieving near-human levels of accuracy without fine-tuning, LLMs have the potential to significantly streamline the digitization process, making vast troves of archival material more accessible to researchers and the public. This has the potential to open-up new possibilities for large-scale textual analysis, information retrieval, and data-driven approaches to the study of history. 

\section{Note on AI Use in this Manuscript}

 GitHub Copilot and Claude Sonnet-3.5 were used to assist with python coding and to generate the documentation for that code; Claude Sonnet-3.5 was used to convert references in the works cited to the correct format; to check the text for errors (specifically: grammar and spelling, missing citation information, formatting, and statistical discrepancies between tables and the text), and to write a first-draft of the abstract.

\section{Funding Details}

This work was supported by the Canada Research Chair Program, (Lianne C. Leddy) and the Insight Grant Program,  (Mark Humphries), both provided by the Social Sciences and Humanities Research Council of Canada.

\section{Disclosure Statement}

There authors report there are no competing interests to declare.

 \section{Works Cited}

Al Kendi, W., F. Gechter, L. Heyberger, and C. Guyeux. 2024. “Advancements and Challenges in Handwritten Text Recognition: A Comprehensive Survey” \textit{Journal of Imaging} 10 (18): 1-30. \url{https://doi.org/10.3390/jimaging10010018}.

Bergen, Teresa. 2019. Transcribing Oral History. New York: Routledge. \url{https://doi.org/10.4324/9781351142007}

Bluche, T. 2016. “Joint Line Segmentation and Transcription for End-to-End Handwritten Paragraph Recognition.” \textit{Advances in Neural Information Processing Systems} 29: 838–846. \url{https://doi.org/10.48550/arXiv.1604.08352}.

Bommasani, R., et al. 2021. “On the Opportunities and Risks of Foundation Models.” \textit{arXiv preprint} 2108.07258 \url{https://doi.org/10.48550/arXiv.2108.07258}

Brown, T., et al. 2020. “Language Models are Few-Shot Learners.” \textit{Advances in Neural Information Processing Systems} 33: 1877–1901. \url{https://dl.acm.org/doi/abs/10.5555/3495724.3495883}

Cheng, Z., F. Bai, Y. Xu, G. Zheng, S. Pu, and S. Zhou. 2017. “Focusing Attention: Towards Accurate Text Recognition in Natural Images.” \textit{Proceedings of the IEEE International Conference on Computer Vision}: 5076–5084. \url{https://doi.ieeecomputersociety.org/10.1109/ICCV.2017.543}

Christlein, V., A. Nicolaou, T. Schlauwitz, S. Späth, K. Herbers, and A. Maier. 2018. “Handwritten Text Recognition Error Rate Reduction in Historical Documents using Naive Transcribers.” INF-DH-2018.\url{https://doi.org/10.18420/infdh2018-13}.

D’Arce, R., T. Norton, S. Hannuna, and N. Cristianini. 2022. “Self-attention Networks for Non-recurrent Handwritten Text Recognition.” \textit{Frontiers in Handwriting Recognition. ICFHR 2022. Lecture Notes in Computer Science} 13639.\url{https://doi.org/10.1007/978-3-031-21648-0_27}

Feng, J.E., A.A. Anoushiravani, P.J. Tesoriero, et al. 2020. “Transcription Error Rates in Retrospective Chart Reviews.” \textit{Orthopedics} 43(5): e404-e408. \url{https://doi.org/10.3928/01477447-20200619-10}

GMR Transcription. 2024. “Transcription Services Pricing.” Accessed October 2, 2024. \url{https://www.gmrtranscription.com/prices\#T2TTranscription}

Google. 2024. “Generating Content.” Google AI for Developers Website. Accessed October 21, 2024. \url{https://ai.google.dev/api/generate-content}

Graves, A., M. Liwicki, S. Fernández, R. Bertolami, H. Bunke, and J. Schmidhuber. 2008. “A Novel Connectionist System for Unconstrained Handwriting Recognition.” \textit{IEEE Transactions on Pattern Analysis and Machine Intelligence} 31(5): 855–868. \url{https://www.cs.toronto.edu/\~graves/tpami\_2009.pdf}

Howard, J., and S. Ruder. 2018. “Universal Language Model Fine-tuning for Text Classification.” \textit{Proceedings of the 56th Annual Meeting of the Association for Computational Linguistics} 1: 328–339. \url{https://doi.org/10.18653/v1/p18-1031}.

Huang, J., X. Chen, S. Mishra, H.S. Zheng, A.W. Yu, X. Song, and D. Zhou. 2024. “Large Language Models Cannot Self-Correct Reasoning Yet.” \textit{ICLR 2024 Conference}. \url{https://openreview.net/pdf?id=IkmD3fKBPQ}.

Humphries, M. “History and Generative AI”. \textit{Teaching History} 57 (3): 4-9. \url{https://search.informit.org/doi/10.3316/informit.322748974380724}.

Humphries, M. and L. Leddy. 2023. “The Future of History”. \textit{Intersections} 6 (3): 21-22. \url{https://cha-shc.ca/wp-content/uploads/2024/05/Intersections-6.3-online.pdf}.

Humphries, M. and E. Story. 2023. “Today’s AI, Tomorrow’s History: Doing History in the Age of ChatGPT,” \textit{Active History}. \url{https://activehistory.ca/2023/03/todays-ai-tomorrows-history-doing-history-in-the-age-of-chatgpt/}.

Kahle, P., S. Colutto, G. Hackl, and S. Gruden. 2017. “Transkribus - A Service Platform for Transcription, Recognition and Retrieval of Historical Documents.” \textit{2017 14th IAPR International Conference on Document Analysis and Recognition }4: 19–24. \url{https://doi.org/10.1109/ICDAR.2017.307}.

Kaplan, J., S. McCandlish, T. Henighan, T.B. Brown, B. Chess, R. Child, S. Gray, A. Radford, J. Wu, and D. Amodei. 2020. “Scaling Laws for Neural Language Models.” arXiv preprint arXiv:2001.08361. \url{https://doi.org/10.48550/arXiv.2001.08361}.

Keeling, A., and J. Sandlos. 2011. “Shooting the Archives: Document Digitization for Historical–Geographical Collaboration.” \textit{History Compass} 9: 423-432. \url{https://doi.org/10.1111/j.1478-0542.2011.00771}.

LeCun, Y., L. Bottou, Y. Bengio, and P. Haffner. 1998. “Gradient-Based Learning Applied to Document Recognition.” \textit{Proceedings of the IEEE} 86(11): 2278–2324. \url{https://doi.org/10.1109/5.726791}.

Leddy, Lianne. 2010. “Interviewing Nookomis and Other Reflections: The Promise of Community Collaboration.” \textit{Oral History Forum} 30: 1-48. \url{https://www.oralhistoryforum.ca/index.php/ohf/article/view/386/457/index.html}.

Liang, X., et al. 2024. “Internal Consistency and Self-Feedback in Large Language Models: A Survey.” arXiv preprint arXiv:2407.14507. \url{https://doi.org/10.48550/arXiv.2407.14507}.

McLean, M. A., Roberts, D. A., and Gibbs, M. 2024. “Ghosts and the machine: testing the use of Artificial Intelligence to deliver historical life course biographies from big data.” \textit{Historical Methods: A Journal of Quantitative and Interdisciplinary History}57: 1–17. \url{https://doi.org/10.1080/01615440.2024.2398455}.

Memon, J., M. Sami, R.A. Khan, and M. Uddin. 2020. “Handwritten Optical Character Recognition (OCR): A Comprehensive Systematic Literature Review (SLR).” \textit{IEEE Access} 8: 142642-142668. \url{https://doi.org/10.1109/ACCESS.2020.3012542}.

Nihart, Nathanael. 2022. “Digital Humanities Techniques for Asylum Studies In the Archive.” Masters Thesis, University of North Carolina. \url{https://doi.org/10.17615/4hnn-kh38}.

Nockels, J., P. Gooding, S. Ames, and M. Terras. 2022. “Understanding the Application of Handwritten Text Recognition Technology in Heritage Contexts: A Systematic Review of Transkribus in Published Research.” \textit{Archival Science} 22(3): 367–392. \url{https://doi.org/10.1007/s10502-022-09397-0}.

Nordo, A.H., E.L. Eisenstein, J. Hawley, S. Vadakkeveedu, M. Pressley, J. Pennock, and I. Sanderson. 2017. “A Comparative Effectiveness Study of eSource Used for Data Capture for a Clinical Research Registry.” \textit{International Journal of Medical Informatics} 103: 89-94. \url{https://doi.org/10.1016/j.ijmedinf.2017.04.015}.

Ó Raghallaigh, Brian. 2022. “Handwritten Text Recognition (HTR) for Irish-Language Folklore.” \textit{Proceedings of the 4th Celtic Language Technology Workshop}: 121-126. \url{https://aclanthology.org/2022.cltw-1.17}.

Oliveira, S.A., and F. Kaplan. 2018. “Comparing Human and Machine Performances in Transcribing 18th Century Handwritten Venetian Script.” \textit{ADHO/EHD 2018-Mexico City}. \url{https://infoscience.epfl.ch/record/255998/files/final\_abstract\_dh18.pdf}.

Pan, L., M. Saxon, W. Xu, D. Nathani, X. Wang, and W.Y. Wang. 2024. “Automatically Correcting Large Language Models: Surveying the Landscape of Diverse Automated Correction Strategies.” \textit{Transactions of the Association for Computational Linguistics} 12: 484–506. \url{https://doi.org/10.1162/tacl\_a\_00660}.

Perdiki, E. 2023. “Preparing Big Manuscript Data for Hierarchical Clustering with Minimal HTR Training.” \textit{Journal of Data Mining and Digital Humanities}. \url{https://doi.org/10.46298/jdmdh.10419}.

Plamondon, R., and S.N. Srihari. 2000. “On-Line and Off-Line Handwriting Recognition: A Comprehensive Survey.” \textit{IEEE Transactions on Pattern Analysis and Machine Intelligence} 22(1): 63–84. \url{https://doi.org/10.1109/34.824821}.

Prebor, Gila. 2023. “From Digitization and Images to Text and Content: Transkribus as a Case Study.” \textit{Proceedings of the Association for Information Science and Technology} 60(1): 1102-1103. \url{https://doi.org/10.1002/pra2.958}.

Romein, C.A., M. Kemman, J.M. Birkholz, J. Baker, M. De Gruijter, A. Meroño-Peñuela, T. Ries, R. Ros, and S. Scagliola. 2020. “State of the Field: Digital History.” \textit{History: The Journal of the Historical Association} 105 (365): 291-312. \url{https://doi.org/10.1111/1468-229X.12969}.

Rumelhart, D.E., G.E. Hinton, and R.J. Williams. 1986. “Learning Representations by Back-propagating Errors.” \textit{Nature} 323(6088): 533–536. \url{https://doi.org/10.1038/323533a0}.

Sánchez, J.A., V. Romero, A.H. Toselli, and E. Vidal. 2019. “ICFHR2018 Competition on Handwritten Text Recognition on the READ Dataset.” \textit{2018 16th International Conference on Frontiers in Handwriting Recognition}: 489–493. \url{https://doi.org/10.1109/ICFHR-2018.2018.00089}.

Sánchez-DelaCruz, E., and C.I. Loeza-Mejía. 2024. “Importance and Challenges of Handwriting Recognition with the Implementation of Machine Learning Techniques: A Survey.” \textit{Applied Intelligence} 54(8): 6444–6465. \url{https://doi.org/10.1007/s10489-024-05487-x}.

Schulhoff, S., et al. 2024. “The Prompt Report: A Systematic Survey of Prompting Techniques.” arXiv preprint arXiv:2406.06608. \url{https://doi.org/10.48550/arXiv.2406.06608}.

Stechly, K., M. Marquez, and S. Kambhampati. 2023. “GPT-4 Doesn't Know It's Wrong: An Analysis of Iterative Prompting for Reasoning Problems.” arXiv preprint arXiv:2310.12397. \url{https://doi.org/10.48550/arXiv.2310.12397}.

Stevens, M. and S. Burg, 1997. \textit{Editing Historical Documents: A Handbook of Practice}. New York: Rowman \& Littlefield.

Stolcke, A., and J. Droppo. 2017. “Comparing Human and Machine Errors in Conversational Speech Transcription.” arXiv preprint arXiv:1708.08615. \url{https://doi.org/10.48550/arXiv.1708.08615}.

Strong, Liz H. 2018. \textit{Oral History Transcription Style Guide}. Columbia University Center for Oral History Research. \url{https://www.utep.edu/liberalarts/oral-history/\_files/docs/ccohr\%20transcript\%20style\%20guide.pdf}.

Tafti, A.P., A. Baghaie, M. Assefi, H.R. Arabnia, Z. Yu, and P. Peissig. 2016. “OCR as a Service: An Experimental Evaluation of Google Docs OCR, Tesseract, ABBYY FineReader, and Transym.” \textit{Advances in Visual Computing. ICVC 2016. Lecture Notes in Computer Science} 10072: 735-746. \url{https://doi.org/10.1007/978-3-319-50835-1\_66}.

Vaswani, A., N. Shazeer, N. Parmar, J. Uszkoreit, L. Jones, A.N. Gomez, Ł. Kaiser, and I. Polosukhin. 2017. “Attention is All You Need.” \textit{Proceedings of the 31st International Conference on Neural Information Processing Systems}: 6000-6010. \url{https://doi.org/10.48550/arXiv.1706.03762}.

Vidal, E., A.H. Toselli, A. Ríos-Vila, and J. Calvo-Zaragoza. 2023. “End-to-end Page-level Assessment of Handwritten Text Recognition.” \textit{Pattern Recognition} 142: 109695. \url{https://doi.org/10.1016/j.patcog.2023.109695}.

Wang, J. 2023. “A Study of The OCR Development History and Directions of Development.” \textit{Highlights in Science, Engineering and Technology} 72: 409-415. \url{https://doi.org/10.54097/bm665j77}.

Wigington, C., C. Tensmeyer, B. Davis, W. Barrett, B. Price, and S. Cohen. 2018. “Start, Follow, Read: End-to-End Full-Page Handwriting Recognition.” \textit{Proceedings of the European Conference on Computer Vision}: 372-388. \url{https://doi.org/10.1007/978-3-030-01231-1\_23}.

Zhou, Y., M. Keuper, and M. Fritz. 2024. “Balancing Diversity and Risk in LLM Sampling: How to Select Your Method and Parameter for Open-Ended Text Generation.” arXiv preprint arXiv:2408.13586. \url{https://doi.org/10.48550/arXiv.2408.13586}.

\section{Data Availability Statement}

The python code used to test error rates (CER\_WER.py) and for the Transcription Pearl software are being made available via an Open Access GitHub Repository at: \url{https://github.com/mhumphries2323/Transcription\_Pearl}.

We are not making the specific document set used for testing publicly available because it would be impossible to prevent it from being used to train Large Language Models, thus rendering it useless for further testing. It should be noted that the actual contents of the data are irrelevant and we expect that testing on a similar corpus of 18\textsuperscript{th} and early 19\textsuperscript{th} century English language texts would produce similar results.

\end{document}